\pdfoutput=1

\documentclass[11pt]{article}

\newcommand\blfootnote[1]{%
  \begingroup
  \renewcommand\thefootnote{}\footnote{#1}%
  \addtocounter{footnote}{-1}%
  \endgroup
}

\usepackage[]{acl}

\usepackage{times}
\usepackage{latexsym}
\usepackage{comment}
\usepackage{color}

\usepackage{colortbl}
\usepackage[T1]{fontenc}

\usepackage[utf8]{inputenc}

\usepackage{microtype}

\usepackage{inconsolata}

\usepackage{mathtools}
\usepackage{makecell}

\newcommand{\alg}{\code{Crayon}\xspace}

\usepackage{graphicx}
\usepackage{amsmath}
\usepackage{amssymb}
\usepackage{booktabs}
\usepackage{algpseudocode}

\usepackage[export]{adjustbox}
\usepackage{float}

\usepackage[capitalize]{cleveref}
\crefname{section}{Sec.}{Secs.}
\Crefname{table}{Table}{Tables}

\usepackage{amsmath,amsfonts,bm}

\def\eqref#1{equation~\ref{#1}}

\def\1{\bm{1}}

\DeclareMathAlphabet{\mathsfit}{\encodingdefault}{\sfdefault}{m}{sl}
\SetMathAlphabet{\mathsfit}{bold}{\encodingdefault}{\sfdefault}{bx}{n}

\usepackage{geometry}
\usepackage{booktabs} 
\usepackage{bbm}
\usepackage{mathtools}
\usepackage{nccmath}
\usepackage{setspace}

\usepackage{caption}
\usepackage{subcaption}

\usepackage[linesnumbered,ruled,vlined]{algorithm2e}

\SetKwInput{KwInput}{Input}                
\SetKwInput{KwOutput}{Output}              

\SetCommentSty{mycommfont}

\SetAlCapSty{algcapsty}

\usepackage[T1]{fontenc}
\usepackage{wrapfig,lipsum,booktabs}

\usepackage{soul}
\usepackage{dsfont}
\usepackage{enumerate}
\usepackage{enumitem}

\usepackage{amsmath}
\usepackage{amsfonts}
\usepackage{bbm}
\usepackage{dsfont}
\usepackage[Symbol]{upgreek}
\usepackage{lscape}
\usepackage{caption}
\usepackage{balance}
\usepackage{xspace}
\usepackage{float}
\usepackage{kotex}

\usepackage{wasysym}
\usepackage{xcolor}
\usepackage{multirow}
\usepackage{array, boldline, rotating}

\usepackage{amssymb}
\usepackage{pifont}
\renewcommand*\eqref[1]{(\ref{#1})}

\newcommand{\yes}[1]{\textcolor{blue}{[YES]}}
\newcommand{\no}[1]{\textcolor{orange}{[NO]}}
\newcommand{\na}[1]{\textcolor{gray}{[N/A]}}
\newcommand\jh[1]{\textcolor{black}{#1}} 
\newcommand{\eg}{\emph{e.g.,~}}
\newcommand{\ie}{\emph{i.e.,~}}

\def\code#1{\texttt{#1}}

\definecolor{LightCyan}{rgb}{0.88,1,1}
\definecolor{Blue}{rgb}{0, 0.5, 1}
\definecolor{Green}{rgb}{0.0, 0.8, 0.0 }
\definecolor{Red}{rgb}{0.95, 0.55, 0.6}
\definecolor{Skyblue}{rgb}{0.6, 0.6, 0.95 }

\NewDocumentEnvironment{suptitle}{ +b }{
    \twocolumn[{#1}]%
}{}

\NewDocumentCommand{\supptitle}{s}{
\begin{suptitle}
        \centering
        \rule{\textwidth}{0.03cm}\\[0.1cm]
        -Supplementary Material-\\[0.2cm]
        {\Large 
            \textbf{\mytitle }
        }\\
        \rule{\textwidth}{0.03cm}\\[0.2cm]
\end{suptitle}}

\newcommand{\llama}{LLaMA}
\newcommand{\tr}{\textrm{tr}}
\newcommand{\per}{\textrm{c}}
\newcommand{\pool}{pool}
\newcommand{\mytitle}{Crayon: Customized On-Device LLM via \\  Instant Adapter Blending and Edge-Server Hybrid Inference}
\title{\mytitle}

\author{Jihwan Bang$^*$\hspace{1em}Juntae Lee$^*$\hspace{1em}Kyuhong Shim\hspace{1em}Seunghan Yang\hspace{1em}Simyung Chang$^\dag$\\
{Qualcomm AI Research$^\ddag$, Qualcomm Korea YH, Seoul, Republic of Korea} \\ 
{\texttt {\small\{jihwbang, juntlee, kshim, seunghan, simychan\}@qti.qualcomm.com}}}

\begin{document}
\maketitle

\blfootnote{\hspace{-1.8em}$^*$ The authors contribute equally. \\$^\dag$ indicates corresponding author.\\$^\ddag$Qualcomm AI Research is an initiative of Qualcomm Technologies, Inc.}
\begin{abstract}

The customization of large language models (LLMs) for user-specified tasks gets important. However, maintaining all the customized LLMs on cloud servers incurs substantial memory and computational overheads, and uploading user data can also lead to privacy concerns. On-device LLMs can offer a promising solution by mitigating these issues. Yet, the performance of on-device LLMs is inherently constrained by the limitations of small-scaled models. To overcome these restrictions, we first propose \alg, a novel approach for on-device LLM customization. \alg begins by constructing a pool of diverse base adapters, and then we instantly blend them into a customized adapter without extra training. In addition, we develop a device-server hybrid inference strategy, which deftly allocates more demanding queries or non-customized tasks to a larger, more capable LLM on a server. This ensures optimal performance without sacrificing the benefits of on-device customization. We carefully craft a novel benchmark from multiple question-answer datasets, and show the efficacy of our method in the LLM customization.

\end{abstract}

\section{Introduction} 

\begin{figure*}[t]
    \centering
    \includegraphics[width=0.92\textwidth]{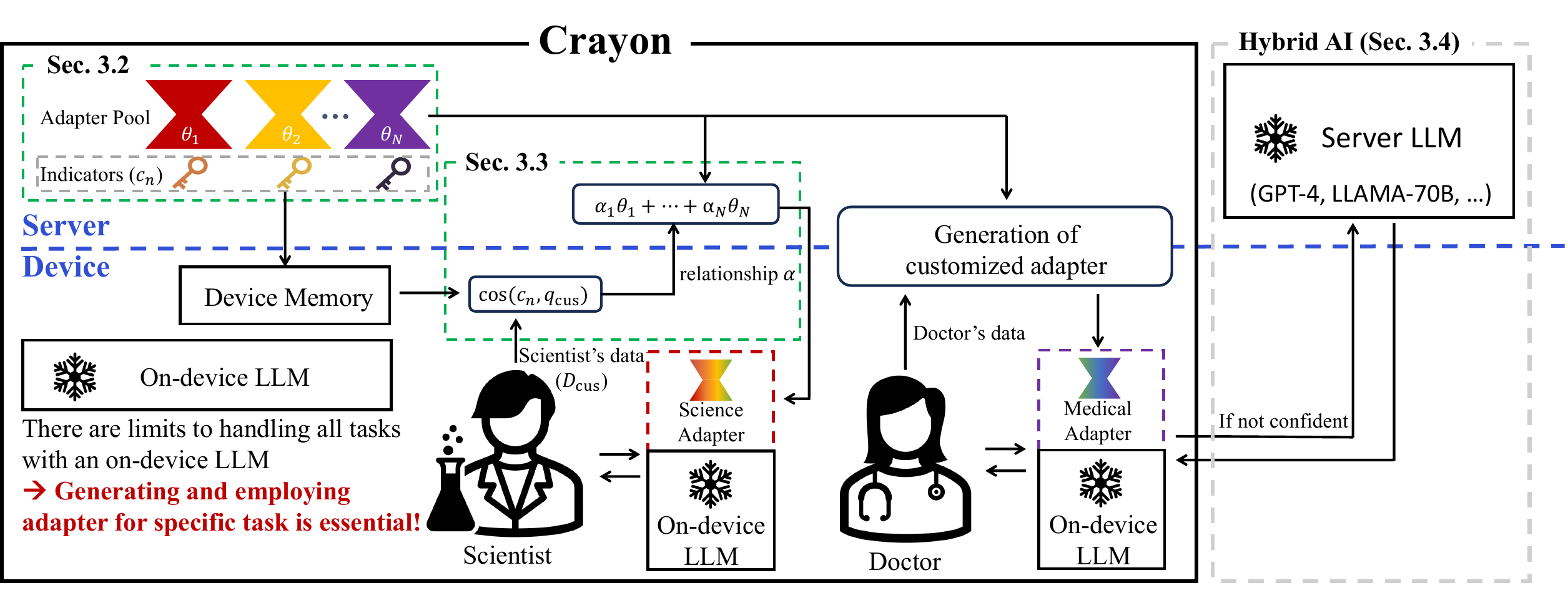}

    \caption{\textbf{Overall framework of the proposed method.} For on-device LLM customization without on-device training cost and privacy issue, we devise \alg generating a suitable adapter instantly by utilizing an adapter pool including preparation of an adapter pool and deploying a customized adapter. Further, we also develop device-server hybrid inference to efficiently leverage a better generalized LLM in the server.}
    \label{fig:intro}
\end{figure*}

Large language model (LLM) has achieved unprecedented success on diverse natural language processing tasks such as machine translation, question and answering, text summarization and stylization, etc. Now then, it is expected for LLM-powered artificial intelligence (AI) to understand and satisfy each user’s unique needs such as recommendation systems, personalized assistance, and personalized search. To this end, a pivotal cornerstone is \textit{customized LLM} where the LLM is highly advanced to a user-requested task. Indeed, several web services related to LLM customization have been emerged such as GPTs~\cite{gpts} and PersonaAI~\cite{characterai, meta_gen}.

However, due to the significant scale of LLM, keeping all the customized LLMs in the servers imposes a tremendous burden. Also, the privacy issues is inevitably raised by uploading the user's data which entail user-requested task. Then, the focus is shifting towards \textit{on-device LLM}. However, as the limited computing power of edge devices, it is impractical to address models as large as those on the servers. Therefore, for on-device LLM customization, it is crucial to maintain the performance on user-defined tasks, while constraining the model sizes. The practical method for the on-device LLM customization, however, has been less explored.

Recently, several works are developed to further lead out the ability of LLMs, and they may be exploited to cover the performance limit of LLMs on smaller size (device-level). \citet{brown2020language} introduced few-shot learning where a few example query-answer prompts are given together with users' queries. Chain-of-thought (COT)~\cite{wei2022chain} tried in-context learning by encouraging LLM to generate evidences as well as final answers. Also, for knowledge-intensive NLP tasks, retrieval augmented generation (RAG)~\cite{lewis2020retrieval} made up query-relevant prompts by retrieving a given database. These prompt-based approaches have a intrinsic problem of increasing inference cost as the prompts get long and complex, and hence they are not suitable for edge devices.

Moreover, we can consider fine-tuning the on-device LLM in order to internalize these prompt-based knowledge for user-specific tasks to the model. Freezing the pre-trained LLM, adapter-based methods~\cite{hu2021lora,houlsby2019parameter,wang2022adamix} have tried to facilitate LLM fine-tuning, and the low-rank-based adapter LoRA~\cite{hu2021lora} have been most in the limelight. Despite on-device-scale LLM armed with the adapters, 
fine-tuning is time-consuming process and also needs a certain-level of training dataset. However, computing power of edge devices is limited and collecting enough user-specific data is also impractical. Thus, we raise inquiry: \textit{How about simply customizing LLM without on-device training?}

For this purpose, given a target customization task, we propose \alg customizing the on-device LLM via a single customized adapter which is blended on-the-fly from a set of base adapters, called \textit{adapter pool}. To cover a wide range of user requests, the base adapters are learned to contain different knowledge each other. 
As depicted in~\autoref{fig:intro}, our approach requires no additional training cost in both of server and edge device when blending the customized adapter. In addition, we develop a device-server hybrid inference strategy to effectively leverage the better-generalized larger model of the server for handling unexpected queries (out-of-scope for the customized model). Our contributions are summarized as follows:
\begin{itemize}
    \item We propose the first practical approach for customization of on-device small-scale LLM.
    \item We develop \alg where the base adapters are learned satisfying their diversification by instantly blending the base adapters, and device-server hybrid inference to cover the out-of-customized tasks. 
    \item We present an on-device LLM customization benchmark by tailoring the public question-answer datasets, and analyze our method.
\end{itemize}

\section{Problem Set-up}

\noindent \textbf{Defining \& processing customized task.} In the context of few-shot learning~\cite{brown2020language}, an LLM is prompted by several query-answer pairs to better understand testing queries. It has been proven that these few-shot prompts are helpful for increasing the generalization capability even in smaller LLMs. 
However, prompting increases the inference cost of LLMs as well, which is not preferred to on-device use case. Rather than prompting the few-shot examples $\mathcal{D}_\per$, we define it as a target task specified to the user. 
From them, we immediately generate an adapter customized for the target task on the server, and deploy it to the on-device LLM. Note that, considering the privacy issue and communication cost with the server, $\mathcal{D}_\per$ itself is never transmitted to the sever in our method.




\noindent\textbf{Baseline LLM.} For autoregressive, causal language model, most of popular LLMs such as GPT~\cite{brown2020language}, \llama~\cite{touvron2023llama}, Mistral~\cite{jiang2023mistral}, and Falcon~\cite{penedo2023refinedweb} have adopted the decoder-only transformer~\cite{vaswani2017attention} architecture. Hence, in our work, we select the smallest \llama~(\llama-7B) as the baseline on-device LLM, which is reasonable size for edge devices\footnote{https://github.com/Bip-Rep/sherpa}.

\noindent\textbf{Adapter for customized LLM.} To reduce training cost, parameter efficient fine-tuning (PEFT) injects small trainable adapters, and only updates them for LLM fine-tuning.
As a widely-used PEFT approach, LoRA~\cite{hu2021lora} approximates
the gradient of pre-trained weights into low-rank matrices, and use them as the LLM adapters.
As such, we also employ LoRA as our LLM adapter. 
Note that the learning LoRA does not take place on edge devices in our method.
Briefly explaining our approach, we only train a set of $N$ base LoRAs $\{l_{\theta_n}\}_{n=1}^{N}$ (\ie LoRA \pool) given a training set $\mathcal{D}_{\tr}$ on the server, and then they are combined and deployed for instant LLM customization to the target task (\ie $\mathcal{D}_{\per}$)  without additional training.

\noindent\textbf{Device-server hybrid inference.} 
Although an on-device LLM is well-customized to a user-specified task, there is inevitable performance gap between the device-level and server-level LLMs. Especially, the on-device LLM suffers from more performance drop when the inputs are out of the target task. Hence, we devise a device-server hybrid inference strategy. When output of an on-device LLM is unconfident, the output is replaced from the server's larger model. To reduce frequent use of the server LLM, we develop a method to determine the reliability of on-device LLM's output inside the device.

\section{Methodology}
\label{sec:method}
In this section, we introduce 
\alg which consists of LoRA pool construction~(\autoref{sec:method_lorapool}) and customized LoRA generation~(\autoref{sec:method_personal_lora}). Also, we develop device-server consistent inference~(\autoref{sec:method_consensus}).

\subsection{Overall Framework of \alg}
\label{sec:method_overall}
As illustrated in \autoref{fig:intro}, given $\mathcal{D}_{\tr}$ that consists of various tasks and a baseline LLM $\mathcal{M}_{\Phi_{0}}$ where $\Phi_{0}$ is the initial weight before customization, we jointly train 
$N$ base LoRAs $\{l_{\theta_n}\}_{n=1}^{N}$ to have different characteristics and knowledge, respectively, in the server. Here, $\theta_n$ is the weight of $l_{\theta_n}$. We also simultaneously learn the base LoRA indicator $c_n$ which is allocated to $l_{\theta_n}$. After training, $N$ base LoRAs (i.e. LoRA \pool{}) and the indicators are located in the server and device, respectively.

Then, for a small-scaled customization dataset $\mathcal{D}_\per$, we first obtain the relationship between a LoRA \pool{} and $\mathcal{D}_\per$ by computing the similarities between the indicators and $\mathcal{D}_\per$ on the device. This similarities are sent to the server, and then used to determine the weights of the base LoRAs in blending the customized LoRA. This customized LoRA is finally deployed to the user's device to customize the on-device baseline LLM to the target task. 
Notice that we only upload the similarities of $\mathcal{D}_\per$ to the indicators, but do not $\mathcal{D}_\per$ itself. This is why our customization is privacy-friendly.

\begin{algorithm}[t]
    \DontPrintSemicolon
    \SetAlgoLined
    \SetNoFillComment
    \caption{Learning LoRA \pool}
    \label{alg:lora_pool}
    \small{\KwInput{
    baseline LLM $\mathcal{M}_{\Phi_0}$, base LoRAs $\{l_{\theta_n}\}_{n=1}^{N}$, training set $\mathcal{D}_{\tr}$}}
    \LinesNotNumbered 
    \small{\textcolor{Skyblue}{\# Extract intermediate embeddings}} \\
    $Q_{\tr} = \{q_x|\mathcal{M}_{\Phi_0}^1(x),  x\in\mathcal{D}_\tr$\} \\
    \small{\textcolor{Skyblue}{\# Set the indicator of each base LoRA}} \\
    $\{c_{n}\}_{n=1}^{N} = \code{K-MEANS\_Centroids}(Q_{\tr}, N)$\\
    \small{\textcolor{Skyblue}{\# Update the base LoRA weights}}\\
    \While{$\mathrm{not\,\,done}$}{
        Compute relationship $\{\alpha_n(q_x)\}_{n=1}^{N}$ (\autoref{eq:relationship})\\
        Compute the combined LoRA weight $\Theta_x$ (\autoref{eq:theta}) \\
        Update $\{\theta_n\}_{n=1}^{N}$ optimizing over $\Theta_x$ (\autoref{eq:objective}) \\
    }
    \KwOutput{Weights of the base LoRAs $\theta_1, \ldots, \theta_N$}
\end{algorithm}

\subsection{\alg: LoRA Pool Construction}
\label{sec:method_lorapool}
To address a variety of target customization tasks, it is important to diversify the base LoRAs' knowledge and characteristics. To this end, we introduce an indicator for each base LoRA. 
In specific, for $\forall$ $x\in\mathcal{D}_\tr$, we first obtain the intermediate embeddings (empirically, the query embeddings of the first self-attention layer) as
\begin{equation}
\label{eq:embed}
    q_x = \mathcal{M}_{\Phi_0}^{1}(x)
\end{equation}
Then, we apply unsupervised $k$-means clustering ($k=N$) with $q_x$ since the text corpora in $\mathcal{D}_\tr$ have no specific task label. To suppress noise and focus on significant features, we reduce the dimensional of the embeddings using PCA before the clustering (see \autoref{sec:app:pca} for more details). For brevity, applying PCA is not explicitly notated.

$N$ centroids $c_n$'s resulting from $k$-means clustering are assigned to each base LoRA, dubbed base LoRA indicators. In the aftermentioned section, the base LoRAs are differently updated depending on the similarity between each corresponding indicator and embedding $q_x$ during training.

\noindent\textbf{Learning base LoRAs.} 
For a training input $x$ in $\mathcal{D}_{\tr}$, we extract its query feature $q_x$ as in the base LoRA indicators. Then, we compute its relationship with the base LoRA $l_{\theta_n}$ by using the corresponding indicator $c_n$:
\begin{equation}
\label{eq:relationship}
    \alpha_n(q_x) = \frac{\mathrm{cos\_sim}(c_n, q_x) + 1}{2}
\end{equation}
where $\mathrm{cos\_sim}(\cdot,\cdot)$ denotes the cosine similarity.

To obtain a LoRA $l_\Theta$ specified to $x$, the $N$ base LoRAs are combined by
\begin{equation}
\label{eq:theta}
    \Theta_x = \alpha_1(q_x)\theta_1 + \alpha_2(q_x)\theta_2 + \cdot\cdot\cdot + \alpha_N(q_x)\theta_N
\end{equation}

Following~\cite{hu2021lora}, we only train the base LoRAs while freezing the baseline LLM. As such, we update the baseline LLM's weights $\Phi_0$ into $\Phi_0+\Delta\Phi(\Theta_x)$ optimizing over $\Theta_x$:
\begin{equation}
\label{eq:objective}
    \max_{\Theta} \sum_{(x,y)\in\mathcal{D}_\tr} \sum_{t=1}^{|y|} \log (p_{\Phi_0+\Delta\Phi(\Theta_x)} (y_t | x, y_{<t}))
\end{equation}
where $y$ is the label for $x$. The entire process is presented in \autoref{alg:lora_pool}.

\begin{algorithm}[t]
    \DontPrintSemicolon
    \SetAlgoLined
    \SetNoFillComment
    \LinesNotNumbered 
    \caption{Generate customized LoRA}
    \label{alg:personalization}
    \small{\KwInput{Base LoRAs $\{l_{\theta_n}\}_{n=1}^{N}$, indicators $\{c_{n}\}_{n=1}^{N}$,  $\mathcal{M}_{\Phi_0}$, a few customized data $\mathcal{D}_\per$
    }}
    \small{\textcolor{Skyblue}{----------------- On-device Processing -----------------}} \\
    \small{\textcolor{Skyblue}{\# Get query embeddings from $\mathcal{D}_\per$ }} \\
    $Q_\per=\{q_{x,\per}|\mathcal{M}_{\Phi_0}^1(x), x\in\mathcal{D}_\per\}$ \\
    \small{\textcolor{Skyblue}{\# Get user embedding}} \\
    $q_{\per} = \frac{1}{|Q_\per|} \sum_{k=1}^{|Q_\per|} q_{x,\per}$, where $q_{x,\per} \in Q_\per$ \\
    \small{\textcolor{Skyblue}{\# Compute relationship between $q_\per$ and $\{l_{\theta_n}\}_{n=1}^{N}$ }}\\
    $\alpha_n^\per=\mathrm{cos\_sim(c_n, q_\per)}, n=1,\ldots,N$ \\
    \small{\textcolor{Skyblue}{------------------- Server Processing -------------------}} \\
    \small{\textcolor{Skyblue}{\# Generate customized LoRA}} \\
    $\hat{\theta} = \Sigma_{i=1}^{N}\alpha_i^\per A_i B_i$ (\autoref{eq:personalized})
    
\KwOutput{customized LoRA $l_{\hat{\theta}}$}
\end{algorithm}

\subsection{\alg: Generation of Customized LoRA}
\label{sec:method_personal_lora}
We explain step-by-step in~\autoref{alg:personalization}. When a user provides a few examples $\mathcal{D}_\per$ describing the customized task, we can generate the customized LoRA instantly from the LoRA pool. To do so, we first obtain an user embedding $q_\per$ to represent the customized task by averaging the query embeddings in $\mathcal{D}_\per$ on the edge device:
\begin{equation}
\label{eq:user_embed}
    q_\per = \frac{1}{|\mathcal{D}_\per|}\sum_{x\in\mathcal{D}_\per}\mathcal{M}_{\Phi_0}^{1}(x)
\end{equation}

After then, following~\autoref{eq:relationship}, we compute the combination ratios $\{\alpha_i^\per\}_{i=1}^{N}$ based on the cosine similarities of $q_\per$ with the base LoRAs indicators $\{c_n\}_{n=1}^{N}$. We upload $\{\alpha_i^\per\}_{i=1}^{N}$  to the server instead of $\mathcal{D}_c$ to generate the customized LoRA, and thus \alg can protect the users' privacy. 
On the server side, since the base LoRA's weight $\theta_i$ can be decomposed by two low-rank trainable weights $A_i$ and $B_i$, we instantly generate the customized LoRA $l_{\hat{\theta}}$:

\begin{equation}
\label{eq:personalized}
    \hat{\theta} = \Sigma_{i=1}^{N}\alpha_i^\per A_i B_i
\end{equation}

 At last, the customized LoRA generated from the LoRA pool is deployed to the edge device, and it customizes the on-device LLM. Accordingly, we can effectively and efficiently customize the on-device LLM without additional training in both server and edge device.

\subsection{Device-Server Hybrid Inference}
\label{sec:method_consensus}
Although the on-device LLM is customized, it cannot accommodate all  kinds of input queries. For instance, the user can raise queries outside the scope of the customized task. Also, even for the queries raised from the customized task, the customized on-device LLM can suffer from its inherent limitation. i.e. relatively small model size. To overcome these difficulties, we intermittently turn to a larger LLM (\eg LLaMA-30B) which is placed on the server due to high computational cost. This server LLM is not customized, but can make better response owing to its superb versatility.



Notice that when deciding if an input query $x$ is routed to the server, we cannot utilize the server LLM $\mathcal{M}_{\Phi_\mathrm{s}}$. Instead, we pre-compute a set of prototypes $S=\{s=\mathcal{M}_{\Phi_\mathrm{s}}(x_\per)|x_\per \in \mathcal{D}_\per\}$ where the server LLM's characteristics is represented. We consider that $S$ is deployed from the server to the edge device, together with $l_{\hat{\theta}}$. 




Then, supposing that the desirable output of the on-device LLM may be close to $S$, we compare the on-device LLM's output $o_x$ with $S$. In specific, we compute the routing score $r_x$ in the edge device:
\begin{equation}
\label{eq:routescore}
    r_x = \frac{1}{|S|}\Sigma_{s\in S}\mathrm{cos\_sim}(o_x, s).
\end{equation}

Finally, we route $x$ to the server LLM when $r_x <r_\mathrm{th}$, or hold the on-device output otherwise. The routing threshold $r_\mathrm{th}$ is empirically determined.
\begin{figure}[t]
    \centering
    \includegraphics[width=0.93\linewidth]{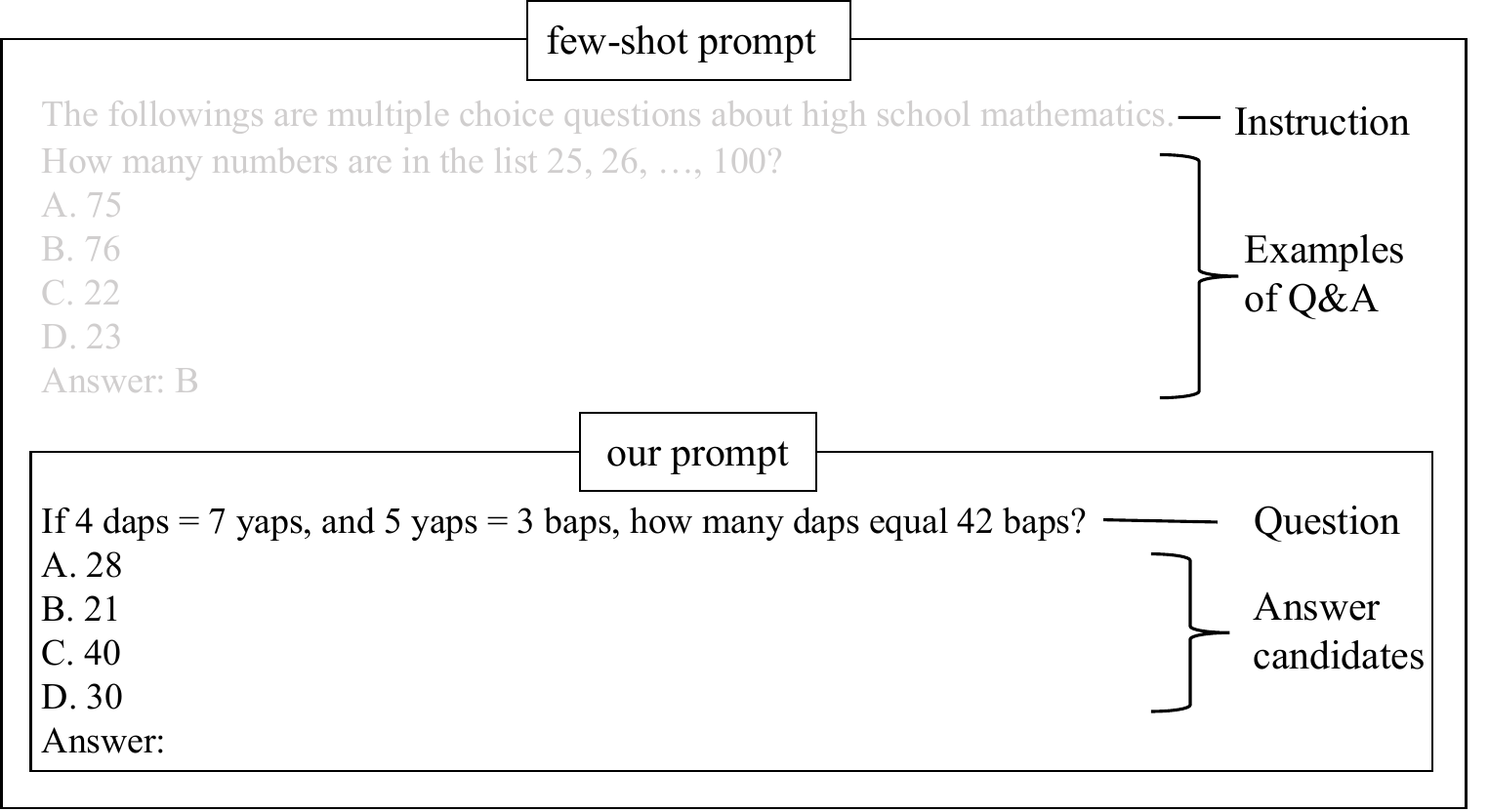}
    \caption{\textbf{Example prompt input in our method.} Different from few-shot prompt, this work does not utilize instruction and examples of QA.}
    \label{fig:prompt_template}
\end{figure}

\begin{table*}[t]
  \centering
  \begin{adjustbox}{width=0.95\linewidth}
  \begin{tabular}{@{}lcccccc@{}}
    \toprule
    \textbf{Method}      &      On-device LLM Size      &               STEM          &      Humanities      &      Social Sciences      &      Other      &      Average  \\ \cmidrule(lr){1-7}
    LLaMA                &        7B            &                 27.8        &         33.2         &           30.9            &      33.0       &       31.0    \\
    LLaMA (few-shot)\textsuperscript{\textdagger}& 7B&            30.5        &         34.0         &           38.3            &      38.1       &       35.1    \\
    LLaMA                &        13B           &                 35.0        &         43.5         &           45.9            &      42.6       &       41.1    \\ 
    LLaMA (few-shot)\textsuperscript{\textdagger}& 13B&           35.8        &         45.0         &           53.8            &      53.3       &       46.9    \\ \cmidrule(lr){1-7}
    Single LoRA         &        7B + 14M            &                  33.2        &         44.6         &           43.4            &      44.6       &       40.7    \\
    Single LoRA (few-shot)      &        7B + 14M            &          29.7        &         33.1         &           33.1            &      39.2       &      33.5 \\
    LoraHub\textsuperscript{\textdaggerdbl}& 7B + 14M&                35.1         &         47.3         &           46.2            &      44.1       &       42.4    \\ \cmidrule(lr){1-7}
    \alg                 &        7B + 14M           &                36.1         &         50.0         &           49.8            &      46.0       &       44.6    \\ 
    \alg + Hybrid(20\%)        &        7B + 14M      &      \textbf{38.6}          &         \textbf{53.6}         &           \textbf{57.6}            &      \textbf{48.6}      &       \textbf{47.6}    \\
   
    \bottomrule
    
    \end{tabular}
    \end{adjustbox}
  \caption{\textbf{Acc (\%) for MMLU tasks.} \textdagger{} indicates the reported performance from original paper~\cite{touvron2023llama} that utilize few-shot learning by following ~\cite{hendrycks2020measuring}. \textdaggerdbl{} mostly follows ~\cite{huang2023lorahub}, but we modify the base model (FLAN-T5 $\xrightarrow{}$ LLaMA-7B) and upstream tasks (BBH $\xrightarrow{}$ \{SIQA, MCQA, OBQA\}). For more details, see \autoref{sec:app:baselines}.
  } 
  \label{tab:mmlu_mainresults}
\end{table*}
\section{Experiment}
\jh{As there is no established benchmark for on-device LLM customization, we present a novel benchmark for this field in~\autoref{ssec:benchmark}. Then, we show comprehensive evaluation and analyses in~\autoref{ssec:main_exp}\&~\autoref{ssec:ablation}. Further details on experimental set-up and baselines in \autoref{sec:app:exp_detail}}.

\subsection{On-device Customization Benchmark}
\label{ssec:benchmark}

\noindent \textbf{Datasets.} 
QA (question-answering) datasets are widely used in evaluation of LLM. Hence, to verify the effectiveness of \alg, we select three public multiple choice QA datasets for training the pool of base LoRAs; \code{Social Interaction QA} (SIQA)~\cite{sap2019social} that focuses on the reasoning about people’s actions and their social implications, \code{MedMCQA} (MCQA)~\cite{pal2022medmcqa} that addresses real-world medical entrance exam questions, and \code{Openbook QA} (OBQA)~\cite{mihaylov2018can} that contains open book exams for assessing human understanding of a subject. For validating task generalization, we utilize \code{MMLU}~\cite{press2022measuring} that contains 57 subjects across STEM, the humanities, the social sciences, and more.

\noindent \textbf{Customized task configuration.} 
In customization datasets, it is expected to contain conversations for different users, where each conversation comprising a series of user's question and LLM's response. However, there are no publicly available datasets. Despite several dialog datasets such as shareGPT~\cite{sharegpt} and ChatAlpaca~\cite{ChatAlpaca}, they lack user identities for the dialogue and then it is difficult to measure the customization results in these datasets. In contrast, the MMLU dataset including individual question-answer data provides annotated by the subject categories. We therefore consider each subject as the interest specified to a user, i.e. a distinct customization task. In specific, we assume 57 distinct users with their own customization tasks. From this, we can quantitatively evaluate customization results in terms of accuracy (Acc). In this experiment, we set the size of customized dataset $|\mathcal{D}_\per|$ as 10 for every user.

\begin{table*}[t!]

  \centering
  \begin{adjustbox}{width=0.98\linewidth}
  \begin{tabular}{@{}lcccccccc@{}}
    \toprule
                              & \multicolumn{8}{c}{Customized Data ($\mathcal{D}_\per$) } \\
                              &              \multicolumn{2}{c}{STEM}                     &                  \multicolumn{2}{c}{Humanities}       &               \multicolumn{2}{c}{Social Science} &         \multicolumn{2}{c}{Others}               \\  
    Task                      &      Elementary Mathematics  &     HS Physics             &        Jurisprudence         &     World Religion     &        HS Geography    & Professional Psychology &      Anatomy             &    Management          \\ \cmidrule(lr){1-1} \cmidrule(lr){2-3} \cmidrule(lr){4-5} \cmidrule(lr){6-7} \cmidrule(lr){8-9}
    
    Elementary Mathematics    &            \textbf{27.2}     &          26.2              &                26.5          &         27.2           &      24.9              &       26.2              &       27.2               &       25.4             \\
    HS Physics                &            28.5              &          \textbf{31.1}     &                29.1          &         31.1           &      31.1              &       31.1              &       26.5               &       30.5             \\
    Jurisprudence             &            54.6              &          54.6              &                \textbf{54.6} &         50.9           &      52.8              &       52.8              &       54.6               &       49.1             \\
    World Religion            &            69.0              &          70.2              &                70.2          &         \textbf{70.2}  &      69.0              &       67.8              &       69.0               &       69.0             \\
    HS Geography              &            58.1              &          56.1              &                57.1          &         59.1           &      \textbf{60.1}     &       55.6              &       59.1               &       59.1             \\
    Professional Psychology   &            40.8              &          41.2              &                41.5          &         40.5           &      38.2              &       \textbf{41.8}     &       41.7               &       36.6             \\
    Anatomy                   &            46.7              &          49.6              &                49.6          &         47.4           &      44.4              &       43.7              &       \textbf{49.6}      &       42.2             \\
    Management                &            52.4              &          51.5              &                50.5          &         57.3           &      57.3              &       52.4              &       52.4               &       \textbf{58.3}     \\
    \bottomrule
    
    \end{tabular}
    \end{adjustbox}
  \caption{\textbf{Ablation study for importance of customization data.} We randomly select two subjects in four categories (\ie STEM, humanities, social science, and others), and the customization performance (Acc) is consistently the highest when the model is customized by using the corresponding dataset.
  } 
  \label{tab:abl_subject}
\end{table*}

\subsection{Main Results}
\label{ssec:main_exp}
\textbf{\alg attains customization.} We evaluate the effectiveness of our method in comparison with several baselines: LLaMA~\cite{touvron2023llama}, LoRA~\cite{hu2021lora}, and LoraHub~\cite{huang2023lorahub}. For LLaMA, we compare the proposed \alg with the larger LLaMA-13B as well as the LLaMA-7B. For fair comparison with our approach, we report both zero and few-shot results. In zero-shot, the input is prompted as in our unified template \autoref{fig:prompt_template}. In few-shot, task-specific few-shot prompt including $\mathcal{D}_\per$ is used following~\cite{touvron2023llama}. In single LoRA, we train LoRA on top of the LLaMA-7B with $\mathcal{D}_\tr$, and it is used universally all the customized tasks. Similar to our method, LoraHub combines pre-trained LoRAs to create a new one suitable for a new task. However, there are several limitations that assume specific upstream tasks with the corresponding datasets to individually pre-train all the LoRAs, and also requires a time-consuming searching process to determine the combination ratios of the pre-trained LoRAs (More details in \autoref{tab:abl:module}).

As shown in \autoref{tab:mmlu_mainresults}, \alg outperforms all the compared methods with only on-device LLM. 
Once the base LoRAs are established in the server, our method enables superior customization (40.7\% vs 44.6\%) compared to Single LoRA with $r=4$ at an identical inference cost in the edge device. Interestingly, \alg  surpasses LoraHub~\cite{huang2023lorahub} by 2.2\% in average, although the base LoRAs are combined instantly. 
It is intuitive that the variety of base LoRAs is beneficial for the generalization ability of LoRA pool. Hence, the outperforming performance of the proposed \alg indicates that our joint training of all the base LoRAs at once produces their diversity, more effectively. This experiment will be addressed in \autoref{ssec:ablation}.


\noindent \textbf{Device-server hybrid inference.}
To assess the efficacy of device-server hybrid inference, we employ LLaMA-30B as the server LLM, which yields 53.2\% Acc on MMLU in average. Moreover, we set the routing threshold $r_\mathrm{th}$ to satisfy 20\% routing ratio, empirically.
As shown in \autoref{tab:mmlu_mainresults}, with only 20\% routing to sever model, the proposed hybrid inference `\alg + hybrid 20\%' obtains 47.6\% Acc, which is even better than fully using the large 13B model. 
Thus, our approach can boost the customized on-device LLM by efficiently intervening a more versatile server LLM.

\subsection{Further Analyses}
\label{ssec:ablation}
We extensively analyze the key components of the proposed method. 

\noindent \textbf{Impact of customized data.}
To assess the efficacy of customized data $\mathcal{D}_{\per}$, we randomly select two different subjects (\ie two customization datasets) from each category in MMLU, and summarize the results in \autoref{tab:abl_subject}. This demonstrates that customized LoRAs, when generated from their corresponding subjects, perform better on their matched subjects than when they are created from unrelated subjects.
 For instance, on the \texttt{Management} subject, the LoRA generated from \texttt{Management} data can obtain 7.8\% higher Acc than the LoRA generated from \texttt{Jurisprudence} data. We can see a similar trend in other subjects.
 Hence, in spite of a very small-scale customization data $\mathcal{D}_{\per}$, it can contain significant generalization cue for the target customization task. Then, unlike the compared methods, our \alg can effectively leverage it for LLM customization.

\begin{figure}[t!]
    \centering
    \includegraphics[width=0.98\linewidth]{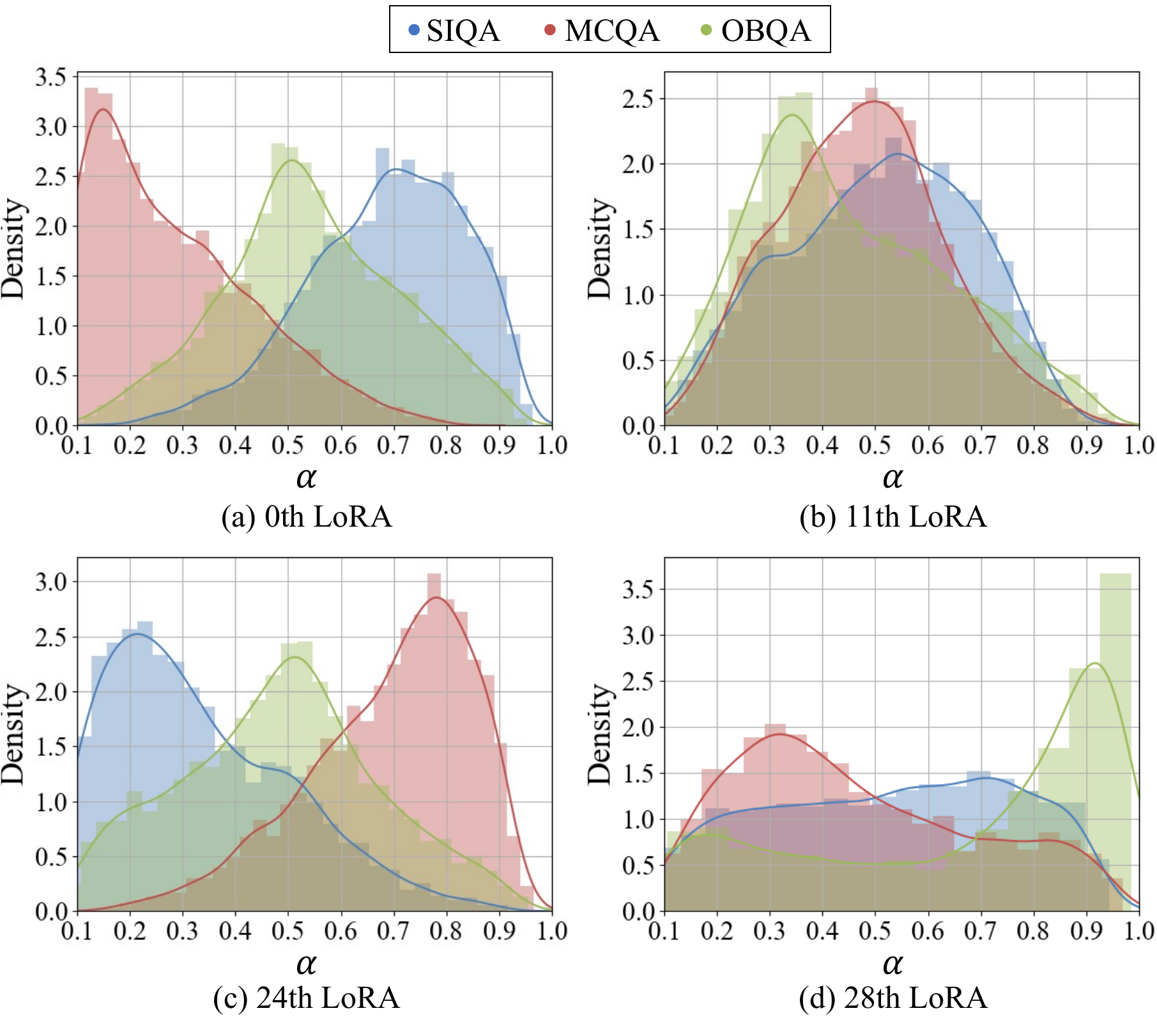}

    \caption{\textbf{Distribution plot of $\alpha$ for each training task on four base LoRAs.} In a, c, and d, the 0th, 24th and 28th base LoRAs have different preference on the SIQA, MCQA, and OBQA tasks, respectively. In b, the 11th base LoRA is trained on all the three tasks evenly.}
    \label{fig:cluster_alpha}
\end{figure}

\noindent \textbf{Diversity in LoRA pool.}
\autoref{fig:cluster_alpha} plots that density distribution~\footnote{It represents the proportion of the data in each range.} of $\alpha$ for each task contained in the training dataset, which shows the focus of the base LoRAs on the training tasks. From 32 base LoRAs, we select the four ones. We can identify that the trained base LoRAs have different weights for each task. 0th base LoRA (\autoref{fig:cluster_alpha}a), 24th base LoRA (\autoref{fig:cluster_alpha}c), and 28th base LoRA (\autoref{fig:cluster_alpha}d) more specialized to  SIQA, MCQA, OBQA, respectively. Unlike above three base LoRAs, 11th base LoRA is likely to evenly trained with all the tasks as shown in \autoref{fig:cluster_alpha}b. Note that we do not provide any information to specify or define the task (e.g. task name) during training. However, \alg produces a diversity of base LoRAs, enabling the LoRA pool to accommodate a wide range of customization tasks.

\begin{figure}[t]
     \centering
     \begin{subfigure}[b]{0.236\textwidth}
         \centering
         \includegraphics[width=0.99\textwidth]{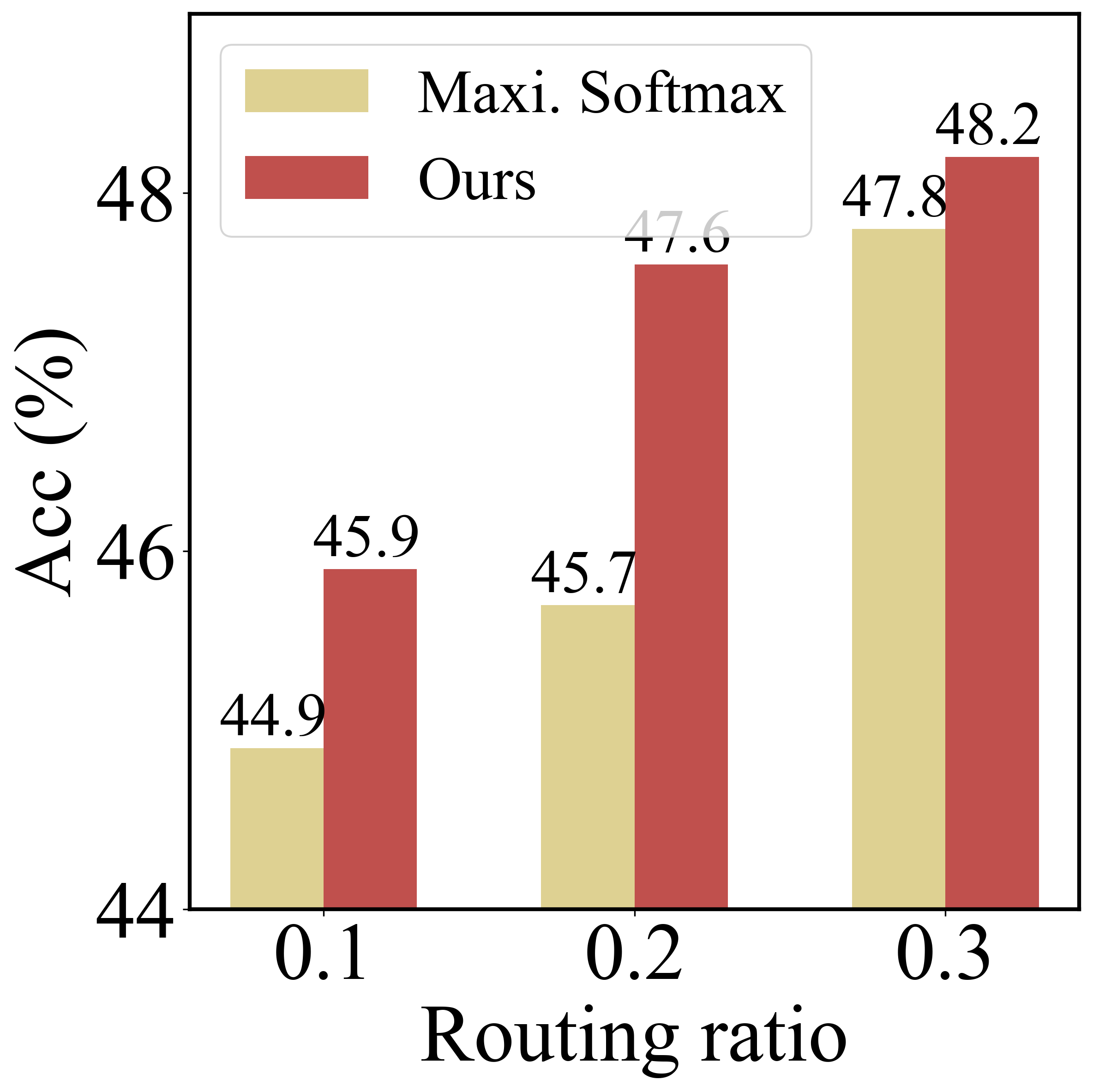}
         \caption{}
         \label{fig:plot_comp_a}
     \end{subfigure}
     \hfill
     \begin{subfigure}[b]{0.236\textwidth}
         \centering
         \includegraphics[width=0.99\textwidth]{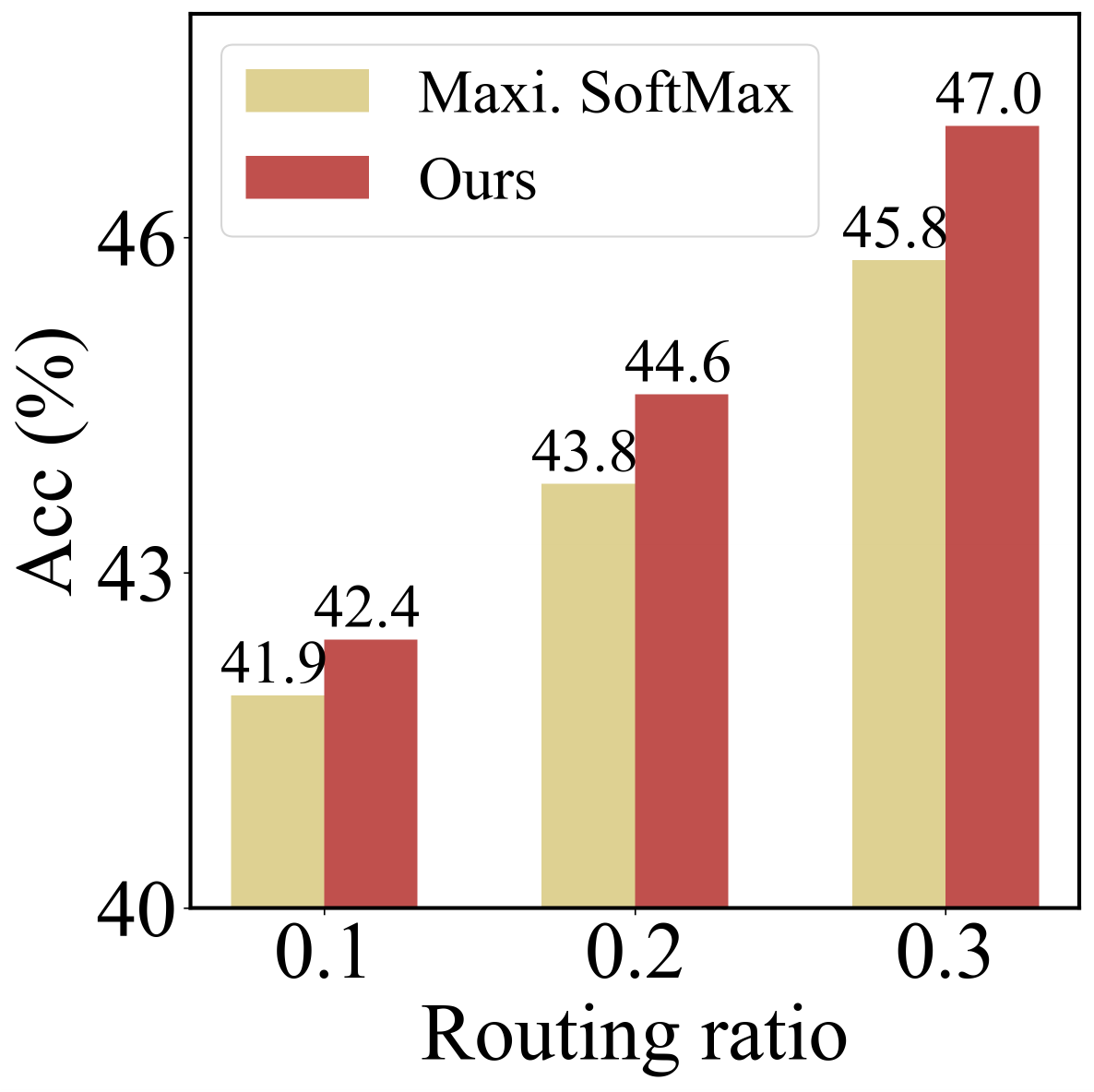}
         \caption{}
         \label{fig:plot_comp_b}
    \end{subfigure}

    \caption{\textbf{Device-server hybrid inference varying routing ratio.}  Acc (\%) on (a) customized tasks and (b) mix of customized \& out-of-customized tasks.}
    
     \label{fig:plot_comp}
\end{figure}

\noindent \textbf{Device-server hybrid inference.}
In deep models, the confidence level is usually estimated by the maximum softmax score, and hence it can be a straightforward choice as decision rule for the hybrid inference. Hence, as in \autoref{fig:plot_comp}, we compare the proposed routing strategy with the maximum softmax-based approach varying the routing ratio. We can see the proposed approach beats the maximum softmax-based approach for all the three routing ratios in \autoref{fig:plot_comp_a}. Hence, our method can effectively detect the failure cases of the customized on-device LLM under a routing ratio. To consider a more general setting, we also assess the capability to address the input queries outside the customized task. To this end, in \autoref{fig:plot_comp_b}, the testing queries are configured from a non-customized subject as well as the customized one. In this setting, we can see a similar trend.
Hence, the our routing effectively complements the customized on-device LLM, and completes the practical use of on-device LLM.

\begin{figure}[t]
     \centering
     \begin{subfigure}[b]{0.236\textwidth}
         \centering
         \includegraphics[width=0.99\textwidth]{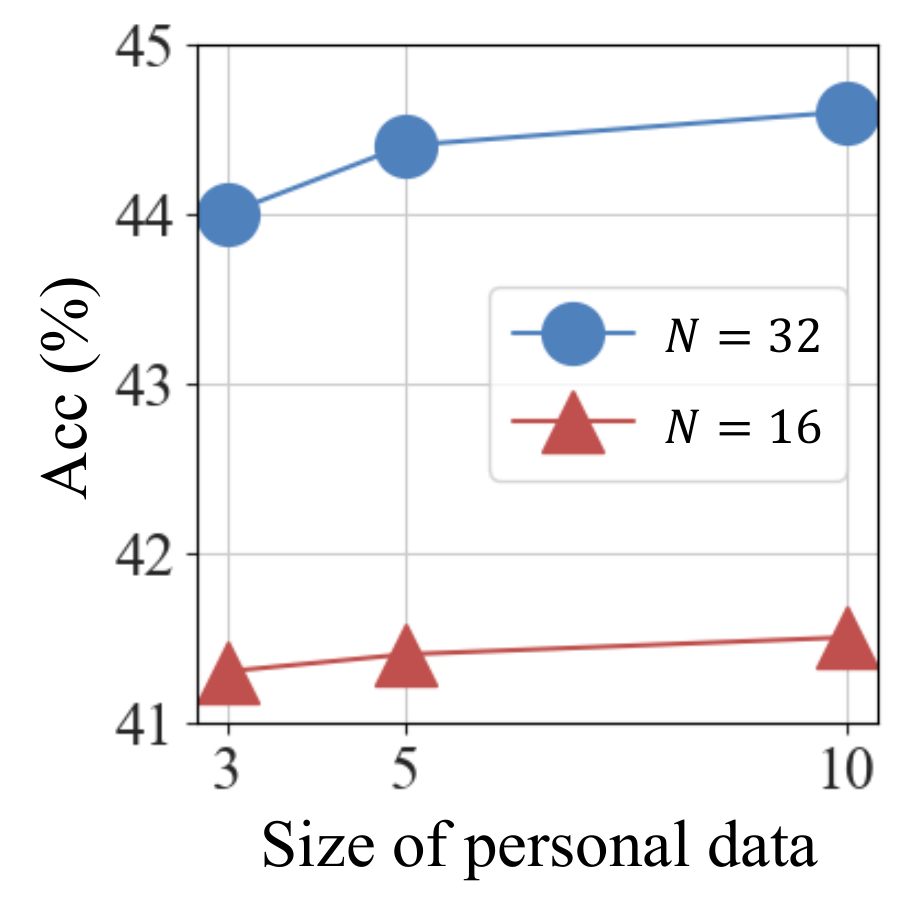}
         \caption{}
         \label{fig:num_user_data}
     \end{subfigure}
     \hfill
     \begin{subfigure}[b]{0.236\textwidth}
         \centering
         \includegraphics[width=0.99\textwidth]{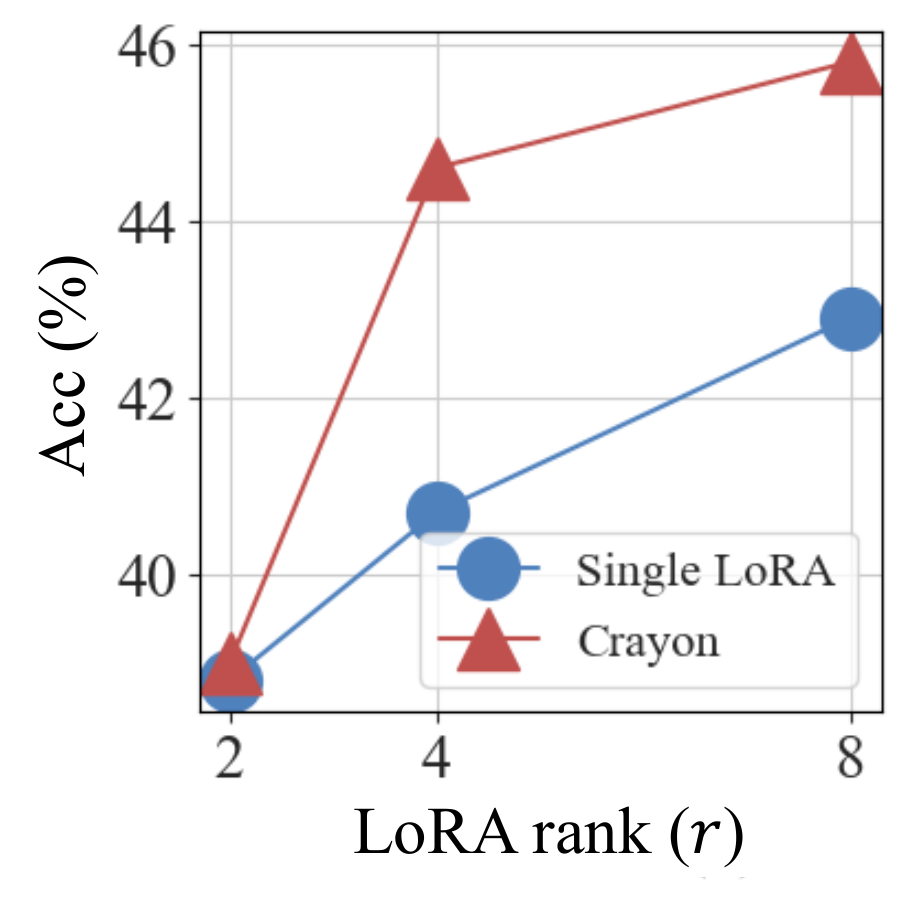}
         \caption{}
         \label{fig:rank}
    \end{subfigure}

    \caption{\textbf{Acc (\%) according to (a) the size of the customized dataset and (b) LoRA rank.}}
    
     \label{fig:ablation}
\end{figure}

\noindent \textbf{Hyperparameter sensitivity.}
\autoref{fig:ablation} examines \alg changing the  size of the customization dataset ($|\mathcal{D}_{\per}|$) and LoRA rank $r$.
In \autoref{fig:num_user_data}, when the number of base LoRAs $N$ is set as 32 (default setting), the proposed method is capable of customization, irrespective of $|\mathcal{D}_{\per}|$. This trend is mirrored when $N$ is reduced to 16. Whereas, despite the same $|\mathcal{D}_{\per}|$, the performance gap between $N=16$ and $N=32$ is notable. Hence, the number of base LoRAs highly impact the quality of customization. As the number of customized data will differ from user to user and $N$ is usually pre-determined in practical use, our method can be effectively applied in a real world scenario.

Further, we investigate the performance difference as varying rank of LoRA in \autoref{fig:rank}. 
When $r=2$, the performance gap is marginal. We infer that the LoRA pool with too small rank might not effectively represent different customization tasks. Nevertheless, it is still slightly better than the a single universal LoRA without any additional training cost. However, once $r\geq4$, the performance gap between them highly increases. Thus, for customization, the proposed task-wise base LoRA blending is more beneficial than generalizing a single LoRA.

\begin{table}[t]
  \centering
  \begin{adjustbox}{width=\linewidth}
  \begin{tabular}{@{}cccccc@{}}
    \toprule
    \multicolumn{2}{c}{Training base LoRAs}     &       \multicolumn{2}{c}{Determining $\alpha$}    &              &            \\ 
    Task-wise        &        Joint             &       Learning-based    &      Cos. sim.          &    Acc (\%)  &    Time (s)   \\ \cmidrule(lr){1-2} \cmidrule(lr){3-4} \cmidrule(lr){5-5} \cmidrule(lr){6-6}
    \checkmark       &                          &           \checkmark    &                         &      42.4    &    49.7     \\
                     &       \checkmark         &           \checkmark    &                         &      44.5    &    51.0        \\
                     &       \checkmark         &                         &       \checkmark        &      44.6    &    0.2      \\
    \bottomrule
    
    \end{tabular}
    \end{adjustbox}
  \caption{\textbf{Ablation analysis.} Each component of \alg is changed to the matched one of LoraHub. (1st and 3rd rows are LoraHub and complete \alg, each.) 
  } 
  \label{tab:abl:module}
  
\end{table}

\noindent \textbf{Component ablation study.}
In \alg, customization consists of two steps; i) constructing multiple base LoRAs and ii) deploying customized LoRA via blending the base LoRAs. First, we jointly learn the base LoRAs with no task definitions from the training datasets, which effectively diversifies the base LoRAs. Second, we obtain the relationship $\alpha$ via simply mapping to the LoRA indicators. LoraHub combines multiple task-specific LoRAs trained on different upstream tasks, and exhaustively search their relationship with a few examples $\mathcal{D}_c$. To validate the efficacy of those components, we change each component to the matched LoraHub's one. As shown in \autoref{tab:abl:module}, the first row where all the components are ablated corresponds to LoraHub. In the second row, we can infer that our task-agnostic joint learning is more beneficial to learn diverse base LoRAs, compared to the LoraHub's individual LoRA learning with task-definition. Also, in the last row, rather than the time-consuming searching of the relationship $\alpha$, the proposed simple LoRA indicator mapping is more proper to the jointly learned base LoRAs, since the base LoRAs are well-aligned with the LoRA indicators during the training.

\section{Related Works}
\noindent \textbf{Task Generalization.} In NLP, generalizing the language models to a wide range of unseen task has been important for its practical use. 
Addressing models with relatively small sizes (under 0.5B), several works have attempted to adapt them with few-shot examples into unseen tasks. However, CrossFit~\cite{ye2021crossfit} has necessity of prompting the task names as hard prefixes.  ReCross~\cite{lin2022unsupervised} alleviated this constraint via exploiting the retrieved training data, and yet requires an additional cost to retrieve task-friendly data. 

Including a way more parameters, LLMs such as GPT~\cite{brown2020language}, LLaMA~\cite{touvron2023llama, touvron2023llama2}, and Gemini~\cite{teamgemini} has shown impressive results to a wide range of queries without specifically trained in the task of queries. However, their model sizes are too large for edge devices. LLaMA and Gemini also released their smaller versions under 10B training parameters which are feasible in edge devices, but their task generalization ability largely lags behind the larger models. To leverage these smaller LLMs, Mistral-7B~\cite{jiang2023mistral} applied sliding window attention and rolling buffer cache. Nevertheless, it cannot cover various tasks as much as the larger models. 
Recently, LoraHub~\cite{huang2023lorahub} pre-trains LoRA adapters for multiple upstream tasks, and generate task-specific LoRA by mixing the pre-trained ones. However, it requires manipulating upstream tasks, and time-consuming process to determine mixing ratio of the pre-trained LoRAs. 
Contrarily, our method constructs the LoRA adapters which has different knowledge and characteristic each other, with no task ques.

\noindent \textbf{Mixture of Experts (MOE).} In MOE~\cite{jacobs1991adaptive, masoudnia2014mixture, riquelme2021scaling}, 
each expert is controlled by a unique gating network, activated based on the distinct nature of the input data. Especially, in language domain, the MoE network identifies and engages the most suitable experts for every token. MoLoRA~\cite{zadouri2023pushing} and SiRA~\cite{zhu2023sira} propose mixture of LoRA, and all the LoRAs and partial LoRAs (\ie top-K) are participating in the gating for every token, respectively. \jh{Moreover, very recently, Mixtral 8x7B~\cite{jiang2024mixtral} has been introduced and surpasses LLaMA-2 70B across all evaluated benchmarks.
It employs eight specialized experts that focus on dense matrices within fully connected layers. During the processing of a token, a routing mechanism selects two of these experts, and their resulting outputs are then merged together.} 
Since edge devices has limited storage to contain several LoRAs, token-wise MoE methods are hard to be applicable in our setup. Several MoE works such as Task-MoE~\cite{kudugunta2021beyond} and Skill Selection~\cite{ponti2023combining} selects experts for every task, and can be adapted to on-device customization. However, they still assumed that the task id should be given with inputs while both training and inference phases.

\noindent \textbf{Speculative Decoding.} To accelerate LLM decoding, speculative decoding~\cite{chen2023accelerating,leviathan2023fast,yang2023inference, gupta2024language} exploits a small (draft) model to predict what the larger target model will produce using uncertainty metric~\cite{fadeeva2023lm}, and then use the target model just to check if the prediction is correct. 
Device-server hybrid inference has a problem that the draft (\ie on-device) model cannot verify the prediction of its own for routing decision without target (\ie server) model. Even though the draft model's output is correct, it should be verified after communication with the server. To solve it, AutoMix~\cite{madaan2023automix} utilized meta-verifier to double-check the self-verification results. To improve both accuracy and efficiency, we develop advanced device-server hybrid inference where the on-device LLM's reliability inside the device. Further, owing to the robust customization via \alg, our customized on-device LLM do not need frequent intervention of the server LLM.


\section{Conclusions}
We propose \alg and device-server hybrid inference for customizing on-device LLM for the first time. In \alg, we jointly train a pool of base LoRAs, which has distinct knowledge and characteristics each other. Using this base LoRA pool, we can instantly blend the base LoRAs into a customized LoRA for user-defined task, without additional training or transferring user data to the server. 
To encompass complex queries, we develop the device-server hybrid system that measures the reliability of the customized LLM to identify the necessity of server LLM.
We also present a new benchmark for quantitative evaluation of on-device LLM customization, incorporating commonly-used QA datasets, and our method shows superior performance. We hope that this benchmark can be a valuable tool for future research in this field.


\section{Limitations} 
\label{sec:limitation}
This work has demonstrated that through an adapter pool elaborately learned from a variety of datasets, it is possible to create customized adapters suitable for unseen target QA tasks. Our methodology is not limited to QA but is also anticipated to be applicable across more NLP domains that requires sequential output tokens. Alongside this, we expect that increasing the number of adapters in the adapter pool together with the utilization of more large-scale datasets will lead to the creation of more diversified customized adapters for a wider scope of unseen tasks.

\section{Potential Risks} 
As with any LLM, the customized on-device LLM's outputs might inadvertently perpetuate biases present in the training data, requiring careful oversight and potential intervention to ensure fairness and ethical use. Moreover, it is crucial to consider the battery life of the edge device of deploying additional computational resources, as the use of edge devices for intricate LLM tasks could result in increased energy consumption.

\bibliography{custom}

\begin{thebibliography}{40}
\expandafter\ifx\csname natexlab\endcsname\relax\def\natexlab#1{#1}\fi

\bibitem[{Bian et~al.(2023)Bian, Lin, Lu, Han, Sun, and He}]{ChatAlpaca}
Ning Bian, Hongyu Lin, Yaojie Lu, Xianpei Han, Le~Sun, and Ben He. 2023.
\newblock Chatalpaca: A multi-turn dialogue corpus based on alpaca instructions.
\newblock \url{https://github.com/cascip/ChatAlpaca}.

\bibitem[{Brown et~al.(2020)Brown, Mann, Ryder, Subbiah, Kaplan, Dhariwal, Neelakantan, Shyam, Sastry, Askell et~al.}]{brown2020language}
Tom Brown, Benjamin Mann, Nick Ryder, Melanie Subbiah, Jared~D Kaplan, Prafulla Dhariwal, Arvind Neelakantan, Pranav Shyam, Girish Sastry, Amanda Askell, et~al. 2020.
\newblock Language models are few-shot learners.
\newblock \emph{Advances in neural information processing systems}, 33:1877--1901.

\bibitem[{Character.AI(2023)}]{characterai}
Character.AI. 2023.
\newblock \url{https://beta.character.ai/}.

\bibitem[{Chen et~al.(2023)Chen, Borgeaud, Irving, Lespiau, Sifre, and Jumper}]{chen2023accelerating}
Charlie Chen, Sebastian Borgeaud, Geoffrey Irving, Jean-Baptiste Lespiau, Laurent Sifre, and John Jumper. 2023.
\newblock Accelerating large language model decoding with speculative sampling.
\newblock \emph{arXiv preprint arXiv:2302.01318}.

\bibitem[{Fadeeva et~al.(2023)Fadeeva, Vashurin, Tsvigun, Vazhentsev, Petrakov, Fedyanin, Vasilev, Goncharova, Panchenko, Panov et~al.}]{fadeeva2023lm}
Ekaterina Fadeeva, Roman Vashurin, Akim Tsvigun, Artem Vazhentsev, Sergey Petrakov, Kirill Fedyanin, Daniil Vasilev, Elizaveta Goncharova, Alexander Panchenko, Maxim Panov, et~al. 2023.
\newblock Lm-polygraph: Uncertainty estimation for language models.
\newblock \emph{arXiv preprint arXiv:2311.07383}.

\bibitem[{Gupta et~al.(2024)Gupta, Narasimhan, Jitkrittum, Rawat, Menon, and Kumar}]{gupta2024language}
Neha Gupta, Harikrishna Narasimhan, Wittawat Jitkrittum, Ankit~Singh Rawat, Aditya~Krishna Menon, and Sanjiv Kumar. 2024.
\newblock Language model cascades: Token-level uncertainty and beyond.
\newblock \emph{arXiv preprint arXiv:2404.10136}.

\bibitem[{Hendrycks et~al.(2020)Hendrycks, Burns, Basart, Zou, Mazeika, Song, and Steinhardt}]{hendrycks2020measuring}
Dan Hendrycks, Collin Burns, Steven Basart, Andy Zou, Mantas Mazeika, Dawn Song, and Jacob Steinhardt. 2020.
\newblock Measuring massive multitask language understanding.
\newblock In \emph{International Conference on Learning Representations}.

\bibitem[{Houlsby et~al.(2019)Houlsby, Giurgiu, Jastrzebski, Morrone, De~Laroussilhe, Gesmundo, Attariyan, and Gelly}]{houlsby2019parameter}
Neil Houlsby, Andrei Giurgiu, Stanislaw Jastrzebski, Bruna Morrone, Quentin De~Laroussilhe, Andrea Gesmundo, Mona Attariyan, and Sylvain Gelly. 2019.
\newblock Parameter-efficient transfer learning for nlp.
\newblock In \emph{International Conference on Machine Learning}, pages 2790--2799.

\bibitem[{Hu et~al.(2021)Hu, Shen, Wallis, Allen-Zhu, Li, Wang, Wang, and Chen}]{hu2021lora}
Edward~J Hu, Yelong Shen, Phillip Wallis, Zeyuan Allen-Zhu, Yuanzhi Li, Shean Wang, Lu~Wang, and Weizhu Chen. 2021.
\newblock Lora: Low-rank adaptation of large language models.
\newblock \emph{International Conference on Learning Representations}.

\bibitem[{Huang et~al.(2023)Huang, Liu, Lin, Pang, Du, and Lin}]{huang2023lorahub}
Chengsong Huang, Qian Liu, Bill~Yuchen Lin, Tianyu Pang, Chao Du, and Min Lin. 2023.
\newblock Lorahub: Efficient cross-task generalization via dynamic lora composition.
\newblock \emph{arXiv preprint arXiv:2307.13269}.

\bibitem[{Jacobs et~al.(1991)Jacobs, Jordan, Nowlan, and Hinton}]{jacobs1991adaptive}
Robert~A Jacobs, Michael~I Jordan, Steven~J Nowlan, and Geoffrey~E Hinton. 1991.
\newblock Adaptive mixtures of local experts.
\newblock \emph{Neural computation}, 3(1):79--87.

\bibitem[{Jiang et~al.(2023)Jiang, Sablayrolles, Mensch, Bamford, Chaplot, Casas, Bressand, Lengyel, Lample, Saulnier et~al.}]{jiang2023mistral}
Albert~Q Jiang, Alexandre Sablayrolles, Arthur Mensch, Chris Bamford, Devendra~Singh Chaplot, Diego de~las Casas, Florian Bressand, Gianna Lengyel, Guillaume Lample, Lucile Saulnier, et~al. 2023.
\newblock Mistral 7b.
\newblock \emph{arXiv preprint arXiv:2310.06825}.

\bibitem[{Jiang et~al.(2024)Jiang, Sablayrolles, Roux, Mensch, Savary, Bamford, Chaplot, Casas, Hanna, Bressand et~al.}]{jiang2024mixtral}
Albert~Q Jiang, Alexandre Sablayrolles, Antoine Roux, Arthur Mensch, Blanche Savary, Chris Bamford, Devendra~Singh Chaplot, Diego de~las Casas, Emma~Bou Hanna, Florian Bressand, et~al. 2024.
\newblock Mixtral of experts.
\newblock \emph{arXiv preprint arXiv:2401.04088}.

\bibitem[{Kudugunta et~al.(2021)Kudugunta, Huang, Bapna, Krikun, Lepikhin, Luong, and Firat}]{kudugunta2021beyond}
Sneha Kudugunta, Yanping Huang, Ankur Bapna, Maxim Krikun, Dmitry Lepikhin, Minh-Thang Luong, and Orhan Firat. 2021.
\newblock Beyond distillation: Task-level mixture-of-experts for efficient inference.
\newblock \emph{arXiv preprint arXiv:2110.03742}.

\bibitem[{Leviathan et~al.(2023)Leviathan, Kalman, and Matias}]{leviathan2023fast}
Yaniv Leviathan, Matan Kalman, and Yossi Matias. 2023.
\newblock Fast inference from transformers via speculative decoding.
\newblock In \emph{International Conference on Machine Learning}, pages 19274--19286. PMLR.

\bibitem[{Lewis et~al.(2020)Lewis, Perez, Piktus, Petroni, Karpukhin, Goyal, K{\"u}ttler, Lewis, Yih, Rockt{\"a}schel et~al.}]{lewis2020retrieval}
Patrick Lewis, Ethan Perez, Aleksandra Piktus, Fabio Petroni, Vladimir Karpukhin, Naman Goyal, Heinrich K{\"u}ttler, Mike Lewis, Wen-tau Yih, Tim Rockt{\"a}schel, et~al. 2020.
\newblock Retrieval-augmented generation for knowledge-intensive nlp tasks.
\newblock \emph{Advances in Neural Information Processing Systems}, 33:9459--9474.

\bibitem[{Lin et~al.(2022)Lin, Tan, Miller, Tian, and Ren}]{lin2022unsupervised}
Bill~Yuchen Lin, Kangmin Tan, Chris Miller, Beiwen Tian, and Xiang Ren. 2022.
\newblock Unsupervised cross-task generalization via retrieval augmentation.
\newblock \emph{Advances in Neural Information Processing Systems}, 35:22003--22017.

\bibitem[{Madaan et~al.(2023)Madaan, Aggarwal, Anand, Potharaju, Mishra, Zhou, Gupta, Rajagopal, Kappaganthu, Yang et~al.}]{madaan2023automix}
Aman Madaan, Pranjal Aggarwal, Ankit Anand, Srividya~Pranavi Potharaju, Swaroop Mishra, Pei Zhou, Aditya Gupta, Dheeraj Rajagopal, Karthik Kappaganthu, Yiming Yang, et~al. 2023.
\newblock Automix: Automatically mixing language models.
\newblock \emph{arXiv preprint arXiv:2310.12963}.

\bibitem[{Mangrulkar et~al.(2022)Mangrulkar, Gugger, Debut, Belkada, Paul, and Bossan}]{peft}
Sourab Mangrulkar, Sylvain Gugger, Lysandre Debut, Younes Belkada, Sayak Paul, and Benjamin Bossan. 2022.
\newblock Peft: State-of-the-art parameter-efficient fine-tuning methods.
\newblock \url{https://github.com/huggingface/peft}.

\bibitem[{Masoudnia and Ebrahimpour(2014)}]{masoudnia2014mixture}
Saeed Masoudnia and Reza Ebrahimpour. 2014.
\newblock Mixture of experts: a literature survey.
\newblock \emph{Artificial Intelligence Review}, 42:275--293.

\bibitem[{Meta(2023)}]{meta_gen}
Meta. 2023.
\newblock \url{https://ai.meta.com/genai/}.

\bibitem[{Mihaylov et~al.(2018)Mihaylov, Clark, Khot, and Sabharwal}]{mihaylov2018can}
Todor Mihaylov, Peter Clark, Tushar Khot, and Ashish Sabharwal. 2018.
\newblock Can a suit of armor conduct electricity? a new dataset for open book question answering.
\newblock In \emph{Proceedings of the 2018 Conference on Empirical Methods in Natural Language Processing}, pages 2381--2391.

\bibitem[{OpenAI(2023)}]{gpts}
OpenAI. 2023.
\newblock \url{https://openai.com/blog/introducing-gpts}.

\bibitem[{Pal et~al.(2022)Pal, Umapathi, and Sankarasubbu}]{pal2022medmcqa}
Ankit Pal, Logesh~Kumar Umapathi, and Malaikannan Sankarasubbu. 2022.
\newblock Medmcqa: A large-scale multi-subject multi-choice dataset for medical domain question answering.
\newblock In \emph{Conference on Health, Inference, and Learning}, pages 248--260. PMLR.

\bibitem[{Penedo et~al.(2023)Penedo, Malartic, Hesslow, Cojocaru, Cappelli, Alobeidli, Pannier, Almazrouei, and Launay}]{penedo2023refinedweb}
Guilherme Penedo, Quentin Malartic, Daniel Hesslow, Ruxandra Cojocaru, Alessandro Cappelli, Hamza Alobeidli, Baptiste Pannier, Ebtesam Almazrouei, and Julien Launay. 2023.
\newblock The refinedweb dataset for falcon llm: outperforming curated corpora with web data, and web data only.
\newblock \emph{arXiv preprint arXiv:2306.01116}.

\bibitem[{Ponti et~al.(2023)Ponti, Sordoni, Bengio, and Reddy}]{ponti2023combining}
Edoardo~Maria Ponti, Alessandro Sordoni, Yoshua Bengio, and Siva Reddy. 2023.
\newblock Combining parameter-efficient modules for task-level generalisation.
\newblock In \emph{Proceedings of the 17th Conference of the European Chapter of the Association for Computational Linguistics}, pages 687--702.

\bibitem[{Press et~al.(2022)Press, Zhang, Min, Schmidt, Smith, and Lewis}]{press2022measuring}
Ofir Press, Muru Zhang, Sewon Min, Ludwig Schmidt, Noah~A Smith, and Mike Lewis. 2022.
\newblock Measuring and narrowing the compositionality gap in language models.
\newblock \emph{arXiv preprint arXiv:2210.03350}.

\bibitem[{Riquelme et~al.(2021)Riquelme, Puigcerver, Mustafa, Neumann, Jenatton, Susano~Pinto, Keysers, and Houlsby}]{riquelme2021scaling}
Carlos Riquelme, Joan Puigcerver, Basil Mustafa, Maxim Neumann, Rodolphe Jenatton, Andr{\'e} Susano~Pinto, Daniel Keysers, and Neil Houlsby. 2021.
\newblock Scaling vision with sparse mixture of experts.
\newblock \emph{Advances in Neural Information Processing Systems}, 34:8583--8595.

\bibitem[{Sap et~al.(2019)Sap, Rashkin, Chen, Le~Bras, and Choi}]{sap2019social}
Maarten Sap, Hannah Rashkin, Derek Chen, Ronan Le~Bras, and Yejin Choi. 2019.
\newblock Social iqa: Commonsense reasoning about social interactions.
\newblock In \emph{Proceedings of the 2019 Conference on Empirical Methods in Natural Language Processing}, pages 4463--4473.

\bibitem[{Team()}]{teamgemini}
Gemini Team.
\newblock Gemini: A family of highly capable multimodal models.
\newblock Technical report, Technical report, Google, 12 2023. URL https://storage. googleapis. com~….

\bibitem[{Tey(2022)}]{sharegpt}
Steven Tey. 2022.
\newblock \url{https://sharegpt.com}.

\bibitem[{Touvron et~al.(2023{\natexlab{a}})Touvron, Lavril, Izacard, Martinet, Lachaux, Lacroix, Rozi{\`e}re, Goyal, Hambro, Azhar et~al.}]{touvron2023llama}
Hugo Touvron, Thibaut Lavril, Gautier Izacard, Xavier Martinet, Marie-Anne Lachaux, Timoth{\'e}e Lacroix, Baptiste Rozi{\`e}re, Naman Goyal, Eric Hambro, Faisal Azhar, et~al. 2023{\natexlab{a}}.
\newblock Llama: Open and efficient foundation language models.
\newblock \emph{arXiv preprint arXiv:2302.13971}.

\bibitem[{Touvron et~al.(2023{\natexlab{b}})Touvron, Martin, Stone, Albert, Almahairi, Babaei, Bashlykov, Batra, Bhargava, Bhosale et~al.}]{touvron2023llama2}
Hugo Touvron, Louis Martin, Kevin Stone, Peter Albert, Amjad Almahairi, Yasmine Babaei, Nikolay Bashlykov, Soumya Batra, Prajjwal Bhargava, Shruti Bhosale, et~al. 2023{\natexlab{b}}.
\newblock Llama 2: Open foundation and fine-tuned chat models.
\newblock \emph{arXiv preprint arXiv:2307.09288}.

\bibitem[{Vaswani et~al.(2017)Vaswani, Shazeer, Parmar, Uszkoreit, Jones, Gomez, Kaiser, and Polosukhin}]{vaswani2017attention}
Ashish Vaswani, Noam Shazeer, Niki Parmar, Jakob Uszkoreit, Llion Jones, Aidan~N Gomez, {\L}ukasz Kaiser, and Illia Polosukhin. 2017.
\newblock Attention is all you need.
\newblock \emph{Advances in neural information processing systems}, 30.

\bibitem[{Wang et~al.(2022)Wang, Agarwal, Mukherjee, Liu, Gao, Awadallah, and Gao}]{wang2022adamix}
Yaqing Wang, Sahaj Agarwal, Subhabrata Mukherjee, Xiaodong Liu, Jing Gao, Ahmed~Hassan Awadallah, and Jianfeng Gao. 2022.
\newblock Adamix: Mixture-of-adaptations for parameter-efficient model tuning.
\newblock \emph{arXiv preprint arXiv:2210.17451}.

\bibitem[{Wei et~al.(2022)Wei, Wang, Schuurmans, Bosma, Xia, Chi, Le, Zhou et~al.}]{wei2022chain}
Jason Wei, Xuezhi Wang, Dale Schuurmans, Maarten Bosma, Fei Xia, Ed~Chi, Quoc~V Le, Denny Zhou, et~al. 2022.
\newblock Chain-of-thought prompting elicits reasoning in large language models.
\newblock \emph{Advances in Neural Information Processing Systems}, 35:24824--24837.

\bibitem[{Yang et~al.(2023)Yang, Ge, Wang, Jiao, Jiang, Yang, Majumder, and Wei}]{yang2023inference}
Nan Yang, Tao Ge, Liang Wang, Binxing Jiao, Daxin Jiang, Linjun Yang, Rangan Majumder, and Furu Wei. 2023.
\newblock Inference with reference: Lossless acceleration of large language models.
\newblock \emph{arXiv preprint arXiv:2304.04487}.

\bibitem[{Ye et~al.(2021)Ye, Lin, and Ren}]{ye2021crossfit}
Qinyuan Ye, Bill~Yuchen Lin, and Xiang Ren. 2021.
\newblock Crossfit: A few-shot learning challenge for cross-task generalization in nlp.
\newblock In \emph{Proceedings of the 2021 Conference on Empirical Methods in Natural Language Processing}, pages 7163--7189.

\bibitem[{Zadouri et~al.(2023)Zadouri, {\"U}st{\"u}n, Ahmadian, Ermi{\c{s}}, Locatelli, and Hooker}]{zadouri2023pushing}
Ted Zadouri, Ahmet {\"U}st{\"u}n, Arash Ahmadian, Beyza Ermi{\c{s}}, Acyr Locatelli, and Sara Hooker. 2023.
\newblock Pushing mixture of experts to the limit: Extremely parameter efficient moe for instruction tuning.
\newblock \emph{arXiv preprint arXiv:2309.05444}.

\bibitem[{Zhu et~al.(2023)Zhu, Wichers, Lin, Wang, Chen, Shu, Lu, Liu, Luo, Chen et~al.}]{zhu2023sira}
Yun Zhu, Nevan Wichers, Chu-Cheng Lin, Xinyi Wang, Tianlong Chen, Lei Shu, Han Lu, Canoee Liu, Liangchen Luo, Jindong Chen, et~al. 2023.
\newblock Sira: Sparse mixture of low rank adaptation.
\newblock \emph{arXiv preprint arXiv:2311.09179}.

\end{thebibliography}

\newpage
\appendix
\supptitle

\section{Details on Experiments}
\label{sec:app:exp_detail}

\subsection{Training setting}
We implemented the proposed and baseline methods based on the Huggingface PEFT library~\cite{peft}. We set the rank $r$ and scaling factor of a LoRA as 4, and the number of base LoRAs as 32. For training, we use the AdamW optimizer with a learning rate 0.0001 which is cosine annealed. We also set the batch size as 128 and the maximum iteration as 800. For all the methods,  we unified the prompt template as shown in \autoref{fig:prompt_template} where task cue is not prompted, which is proper to practical use. All the proposed and baseline methods are implemented with PyTorch 2.0.1 and executed on a single NVIDIA A5000 GPU.

\subsection{Baselines}
Since on-device LLM customization is understudied, we carefully selected three baselines to validate the efficacy of \alg. 

\begin{enumerate}
    \item \textbf{LLaMA}~\cite{touvron2023llama, touvron2023llama2} released publicly available models, and also reported the score on the MMLU dataset. However, the reported scores are obtained using the few-shot prompt as in the upper part of~\autoref{fig:prompt_template}, where both the subject name (i.e., the task name) and examples of the subject are given. This few-shot prompt is not applicable to our on-device customization. For fairness, as well as the score from the literature, we also disclose scores using zero-shot prompt of~\autoref{fig:prompt_template}.
    \item \textbf{Single LoRA} follows the training recipe in the literature~\cite{hu2021lora}. Since the Single LoRA is trained using the entire training dataset $\mathcal{D}_{\tr}$, the same LoRA is used for all the customized tasks. Additionally, for a fair comparison, we fine-tune the Single LoRA (named as "Single LoRA (few-shot)"), which is originally trained on the training dataset, using a few number of examples from customized task $\mathcal{D}_\per$. We observed that this additional fine-tuning yields severe performance drop, and it can be attributed to the insufficient size of $\mathcal{D}_\per$ for customizing the LoRA to the specific task.
    \item \textbf{LoraHub}~\cite{huang2023lorahub} did not focus on tailoring their work for on-device LLM customization, but it can offer a proper baseline for validating \alg. It outlines how to generate LoRAs specified to a new task by using a given few examples and LoRAs trained for other upstream tasks. However, the downside is the lengthy process required to generate new LoRAs due to the reliance on few-shot learning with the new task's examples. Moreover, due to necessities that all upstream tasks should be clearly defined (\ie a meticulously refined dataset is needed), it cannot be seamlessly integrated into on-device LLM customization scenarios.
\end{enumerate}

\section{Details of \autoref{tab:mmlu_mainresults}}
\label{sec:app:baselines}

\autoref{tab:abl:specific_mmlu_results} extends the results from \autoref{tab:mmlu_mainresults} to show the accuracy for each of the 57 subjects in the MMLU dataset. This allows us to see which subjects fall under each category (\ie STEM, Humanities, Social Science, and Other). Additionally, methodologies with a higher average accuracy also tend to yield higher accuracy across individual subjects.

\begin{table*}[t]
  \centering
  \begin{adjustbox}{width=0.99\linewidth}
  \begin{tabular}{@{}lrrrrrrrr|rr@{}}
    \toprule
                &       &       \multicolumn{2}{c}{LLaMA-7B}      &              \multicolumn{2}{c}{LLaMA-13B}          &      \multicolumn{2}{c}{Single LoRA}      &   \multirow{2}{*}{LoRA-Hub}   & \multirow{2}{*}{\alg} & \alg\\
                &       &   zero-shot & few-shot                  &     zero-shot & few-shot               &       zero-shot & few-shot               &  &  & + Hybrid(20\%) \\ \midrule 
Abstract Algebra                    & STEM          &     28.0     &  29.0     &    29.0     &    34.0    &     32.0      &     29.0      &    32.0     &   30.0     &            34.0         \\
Anatomy          & Other           &     32.6     &  37.0     &    40.7     &    45.9    &     48.9      &     41.5      &    52.6     &   49.6     &            50.4         \\
Astronomy        & STEM           &     39.5     &  33.6     &    44.7     &    46.1    &     46.7      &     42.1      &    44.7     &   43.4     &            50.0         \\
Business Ethics  & Other           &     31.0     &  40.0     &    39.0     &    45.0    &     33.0      &     44.0      &    40.0     &   36.0     &            43.0         \\
Clinical Knowledge & Other           &     34.7     &  35.1     &    44.5     &    45.7    &     46.4      &     41.5      &    43.8     &   46.0     &            49.2         \\
College Biology & STEM           &     30.6     &  37.5     &    41.7     &    45.1    &     41.7      &     31.2      &    46.5     &   47.9     &            49.7         \\
College Chemistry & STEM           &     23.0     &  32.0     &    38.0     &    30.0    &     31.0      &     35.0      &    34.0     &   33.0     &            31.0         \\
College Computer Science            & STEM           &     28.0     &  29.0     &    36.0     &    39.0    &     28.0      &     16.0      &    28.0     &   40.0     &           39.0         \\ 
College Mathematics & STEM           &     30.0     &  33.0     &    33.0     &    32.0    &     24.0      &     24.0      &    30.0     &   27.0     &            25.0         \\
College Medicine & Other           &     30.1     &  30.6     &    38.7     &    42.8    &     38.7      &     29.5      &    34.7     &   39.9     &            41.8         \\
College Physics & STEM           &     17.6     &  26.5     &    19.6     &    18.6    &     22.5      &     20.6      &    24.5     &   23.5     &            22.0         \\
Computer Security & STEM           &     33.0     &  45.0     &    56.0     &    65.0    &     45.0      &     39.0      &    49.0     &   47.0     &            55.0         \\
Conceptual Physics & STEM           &     29.4     &  36.6     &    37.4     &    41.3    &     38.3      &     36.6      &    35.3     &   34.0     &            35.1         \\
Econometrics  & Social Science           &     21.1     &  23.7     &    30.7     &    27.2    &     22.8      &     29.8      &    31.6     &   21.1     &            23.2         \\
Electrical Engineering & STEM           &     21.4     &  26.9     &    33.8     &    40.7    &     35.9      &     30.3      &    37.2     &   42.8     &            46.2         \\
Elementary Mathematics & STEM           &     24.3     &  24.3     &    27.8     &    24.9    &     26.5      &     25.7      &    25.7     &   27.2     &            28.8         \\
Formal Logic & Humanities           &     30.2     &  27.0     &    36.5     &    33.3    &     29.4      &     30.2      &    29.4     &   27.0     &            30.2         \\
Global Facts & Other           &     30.0     &  29.0     &    30.0     &    35.0    &     31.0      &     33.0      &    30.0     &   32.0     &            32.0         \\
High School Biology & STEM           &     34.8     &  34.5     &    42.9     &    52.6    &     49.7      &     34.8      &    45.8     &   51.3     &            55.9         \\
High School Chemistry  & STEM           &     29.6     &  28.1     &    31.0     &    28.6    &     34.0      &     30.5      &    37.4     &   41.9     &            44.6         \\
High School Computer Science  & STEM           &     28.0     &  31.0     &    42.0     &    48.0    &     31.0      &     25.0      &    39.0     &   41.0     &            42.0         \\
High School European History  & Humanities           &     33.3     &  44.2     &    45.5     &    61.8    &     47.9      &     25.5      &    52.7     &   59.4     &           63.6         \\ 
High School Geography  & Social Science           &     31.3     &  34.3     &    53.0     &    54.6    &     56.6      &     33.8      &    51.5     &   60.1     &            66.2         \\
High School Government And Politics & Social Science           &     28.5     &  44.6     &    62.2     &    66.3    &     53.9      &     36.3      &    56.5     &   63.2     & 67.2         \\           
High School Macroeconomics  & Social Science           &     27.2     &  35.4     &    38.2     &    44.4    &     37.4      &     30.3      &    36.9     &   41.5     &         43.9         \\   
High School Mathematics  & STEM           &     28.9     &  24.8     &    27.0     &    23.7    &     21.5      &     25.9      &    28.9     &   24.1     &            23.9         \\
High School Microeconomics  & Social Science           &     25.6     &  31.9     &    35.7     &    47.5    &     34.9      &     31.5      &    37.4     &   42.4     &         45.0         \\   
High School Physics  & STEM           &     24.5     &  26.5     &    30.5     &    28.5    &     27.2      &     23.8      &    28.5     &   31.1     &            31.8         \\
High School Psychology  & Social Science           &     28.3     &  47.3     &    52.3     &    60.9    &     53.8      &     36.1      &    58.5     &   62.4     &            65.7         \\
High School Statistics  & STEM           &     22.7     &  35.2     &    30.6     &    30.1    &     32.9      &     25.5      &    33.8     &   29.2     &            34.3         \\
High School US History  & Humanities           &     32.8     &  39.7     &    45.1     &    58.3    &     46.6      &     28.4      &    54.9     &   54.4     &            60.3         \\
High School World History  & Humanities           &     28.7     &  40.9     &    31.6     &    66.2    &     51.9      &     28.7      &    59.1     &   59.9     &            64.8         \\
Human Aging  & Other           &     30.5     &  40.8     &    34.1     &    54.7    &     41.7      &     42.2      &    46.2     &   42.2     &            47.5         \\
Human Sexuality  & Social Science           &     30.5     &  36.6     &    40.5     &    58.8    &     44.3      &     29.0      &    50.4     &   51.9     &            58.8         \\
International Law  & Humanities           &     42.1     &  51.2     &    52.9     &    62.8    &     57.0      &     42.1      &    58.7     &   55.4     &            62.8         \\
Jurisprudence  & Humanities           &     33.3     &  38.9     &    50.0     &    51.9    &     46.3      &     37.0      &    46.3     &   54.6     &            56.7         \\
Logical Fallacies  & Humanities           &     29.4     &  39.3     &    49.1     &    52.8    &     46.6      &     38.0      &    51.5     &   55.8     &            62.3         \\
Machine Learning  & STEM           &     27.7     &  23.2     &    28.6     &    31.3    &     29.5      &     39.3      &    32.1     &   34.8     &            34.8         \\
Management  & Other           &     39.8     &  35.0     &    44.7     &    66.0    &     56.3      &     38.8      &    51.5     &   58.3     &            61.5         \\
Marketing  & Other           &     33.8     &  46.6     &    64.1     &    71.8    &     62.8      &     50.0      &    63.2     &   69.2     &            72.6         \\
Medical Genetics  & Other           &     36.0     &  43.0     &    43.0     &    52.0    &     52.0      &     45.0      &    48.0     &   52.0     &            54.0         \\
Miscellaneous  & Other           &     36.7     &  42.4     &    56.4     &    65.4    &     60.3      &     46.0      &    60.9     &   62.3     &            66.5         \\
Moral Disputes  & Humanities           &     29.5     &  40.2     &    41.9     &    50.9    &     35.0      &     33.8      &    42.5     &   48.8     &            52.5         \\
Moral Scenarios  & Humanities           &     22.7     &  24.3     &    24.6     &    30.1    &     24.0      &     23.9      &    24.7     &   24.0     &            23.5         \\
Nutrition  & Other           &     36.6     &  37.6     &    46.7     &    51.6    &     45.8      &     38.9      &    41.8     &   50.0     &            52.2         \\
Philosophy  & Humanities           &     36.0     &  39.9     &    45.3     &    54.0    &     48.9      &     35.4      &    46.6     &   51.8     &            57.2         \\
Prehistory  & Humanities           &     39.2     &  36.1     &    44.1     &    51.5    &     51.9      &     34.6      &    50.0     &   52.8     &            55.8         \\
Professional Accounting  & Other           &     24.8     &  25.9     &    37.6     &    35.8    &     29.8      &     28.7      &    32.3     &   34.4     &            34.0         \\
Professional Law  & Humanities           &     28.1     &  30.2     &    34.9     &    38.0    &     33.4      &     27.5      &    33.9     &   35.9     &            36.6         \\
Professional Medicine  & Other           &     30.5     &  44.5     &    46.7     &    50.4    &     39.0      &     32.4      &    37.1     &   33.8     &            36.4         \\
Professional Psychology  & Social Science           &     28.3     &  35.1     &    41.0     &    47.7    &     37.6      &     32.5      &    39.5     &   41.8     &            46.7         \\
Public Relations  & Social Science           &     35.5     &  40.9     &    43.6     &    60.9    &     42.7      &     39.1      &    48.2     &   43.6     &            48.2         \\
Security Studies  & Social Science           &     29.8     &  31.8     &    42.9     &    53.9    &     31.4      &     25.7      &    31.4     &   38.8     &            41.5         \\
Sociology  & Social Science           &     35.8     &  46.8     &    52.7     &    61.2    &     45.3      &     35.3      &    54.2     &   63.2     &            68.2         \\
US Foreign Policy  & Social Science           &     49.0     &  46.0     &    58.0     &    80.0    &     60.0      &     38.0      &    58.0     &   68.0     &            74.0         \\
Virology & Other           &     34.3     &  30.1     &    29.5     &    43.4    &     38.0      &     37.3      &    34.9     &   38.6     &            41.5         \\
World Religions & Humanities           &     45.6     &  50.9     &    63.7     &    67.8    &     60.2      &     45.6      &    64.3     &   70.2     &            70.8         \\ \midrule
STEM &                                     &     27.8     &  34.0     &    35.0     &    45.0    &     33.2      &     29.7      &    35.1     &   36.1     &            50.3         \\
Humanities  &                              &     33.1     &  30.5     &    43.5     &    35.8    &     44.5      &     33.1      &    47.3     &   50.0     &            50.0         \\
Social Science  &                          &     30.9     &  38.3     &    45.9     &    53.8    &     43.4      &     33.1      &    46.2     &   49.8     &            42.4         \\
Other            &                          &     33.0     &  38.1     &    42.6     &    53.3    &     44.6      &     39.2      &    44.1     &   46.0     &            46.1         \\ \midrule
Average          &                         &     31.0     &  35.1     &    41.1     &    46.9    &     40.7      &     33.5      &    42.4     &   44.6     &        47.6         \\
    \bottomrule
    
    \end{tabular}
    \end{adjustbox}
  \caption{\textbf{Detailed results of \autoref{tab:mmlu_mainresults} on MMLU.}
  } 
  \label{tab:abl:specific_mmlu_results}
\end{table*}
\section{Effectiveness of PCA}
\label{sec:app:pca}

\begin{figure}[t]
    \centering
    \includegraphics[width=0.98\linewidth]{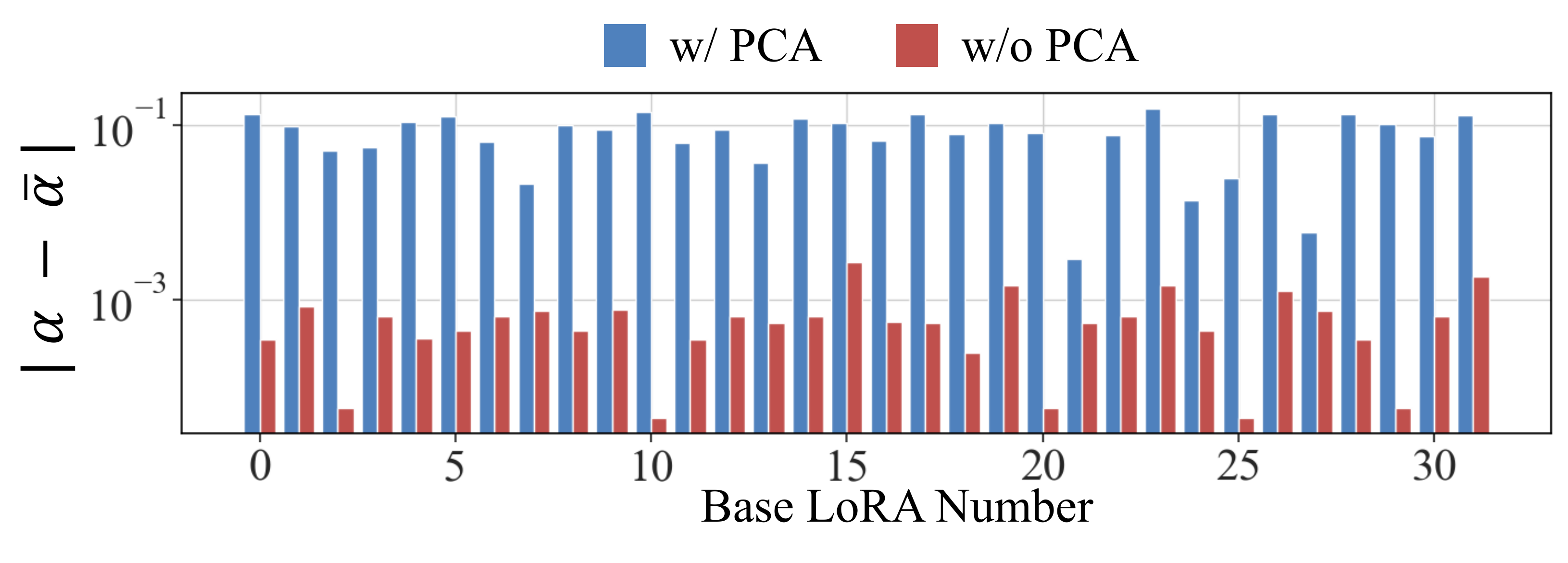}
    \caption{Difference of $\alpha$ and mean of $\alpha$ (\ie $\bar{\alpha}$) from a data point in training set for each base LoRA whether when using PCA or not.}
    \label{fig:pca}
\end{figure}

\autoref{fig:pca} illustrates the deviation of the relationship $\alpha$ corresponding to the base LoRAs for one example in the training set, both with and without the use of PCA. The deviation of $\alpha$ when using PCA is larger 100 times than when not using PCA, implying that we can train the base LoRAs more diversely as experts. In line with the one of the objectives of our method, which is to train the base LoRAs with different types of knowledge, we employ PCA to our methodology when getting embeddings.

\end{document}